\documentclass[conference]{IEEE-conference-template/IEEEtran}
\IEEEoverridecommandlockouts

\usepackage{cite}
\usepackage{amsmath,amssymb,amsfonts}
\usepackage{algorithmic}
\usepackage{graphicx}
\usepackage{textcomp}
\usepackage{xcolor}
\def\BibTeX{{\rm B\kern-.05em{\sc i\kern-.025em b}\kern-.08em
    T\kern-.1667em\lower.7ex\hbox{E}\kern-.125emX}}
\input{definitions}
    
\begin{document}

\title{
    \textit{Sequence-Aware Inline Measurement Attribution for Good-Bad Wafer Diagnosis}\\
    {\huge DM: Big Data Management and Machine Learning}
}
\author{\IEEEauthorblockN{Kohei Miyaguchi$^*$\thanks{$^*$ Kohei Miyaguchi is currently affiliated with LY Research, Japan.}}
\IEEEauthorblockA{\textit{IBM Research -- Tokyo} \\
Tokyo, Japan \\
koheimiyaguchi@gmail.com}
\and
\IEEEauthorblockN{Masao Joko}
\IEEEauthorblockA{\textit{IBM Semiconductors} \\
Tokyo, Japan \\
mjoko@jp.ibm.com}
\and
\IEEEauthorblockN{Rebekah Sheraw}
\IEEEauthorblockA{\textit{IBM Semiconductors} \\
Albany, NY, USA \\
rebekah.sheraw@ibm.com}
\and
\IEEEauthorblockN{Tsuyoshi Id\'e}
\IEEEauthorblockA{\textit{IBM Semiconductors} \\
Yorktown Heights, NY, USA \\
tide@us.ibm.com}
}
\maketitle

\begin{abstract}
How can we identify problematic upstream processes when a certain type of wafer defect starts appearing at a quality checkpoint? Given the complexity of modern semiconductor manufacturing, which involves thousands of process steps, cross-process root cause analysis for wafer defects has been considered highly challenging. This paper proposes a novel framework called Trajectory Shapley Attribution (TSA), an extension of Shapley values (SV), a widely used attribution algorithm in explainable artificial intelligence research. TSA overcomes key limitations of standard SV, including its disregard for the sequential nature of manufacturing processes and its reliance on an arbitrarily chosen reference point. We applied TSA to a good-bad wafer diagnosis task in experimental front-end-of-line processes at the NY CREATES Albany NanoTech fab, aiming to identify measurement items (serving as proxies for process parameters) most relevant to abnormal defect occurrence.
\end{abstract}

\section{Introduction}

Root cause analysis (RCA) of wafer defects is a key challenge throughout all stages of semiconductor manufacturing, from process integration to high-volume production. Performing RCA across multiple process steps is particularly difficult due to the complex sequence and combination of physically and chemically diverse processes, which are carried out using various equipment across hierarchical levels such as wafer, lot, and batch.

RCA is often approached as a by-product of virtual metrology (VM) modeling, which aims to predict a quality metric as a function of process variables. Typically, VM is formulated as a regression or classification problem. However, existing approaches still face challenges that hinder their practicality. Regularized linear regression combined with variable selection techniques, a commonly adopted VM approach (e.g.,~\cite{susto2015multi,jebri2016virtual,kim2018variable}), essentially performs variable-wise correlation analysis for attribution, disregarding the sequential nature of the processing route. While recurrent neural networks (RNNs) and Transformers can capture complex nonlinear dependencies in sequential processes (e.g.,~\cite{yella2021soft,han2023deep,dalla2023deep,lee2020recurrent,hsu2023virtual}), they operate as black boxes, making input attribution a non-trivial task.

\begin{figure}[tbh]
    \centering
    \includegraphics[trim = 60mm 15mm 60mm 95mm, clip,width=1\linewidth]{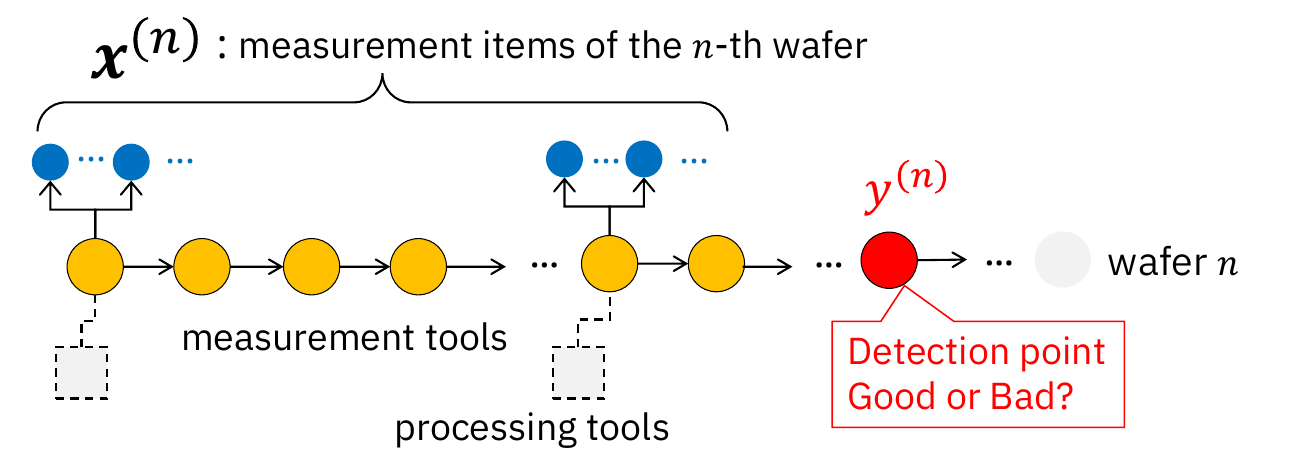}
    \caption{Defect attribution. Given a defect flag $y$ and a set of observed in-line measurements $\bmx$, compute the responsibility score $s_i(\bmx)$ for each measurement item as a proxy for the underlying process.}
    \label{fig:Trajectory_shapley}
\end{figure}

To address the interpretability challenges of deep-learning-based models and facilitate actionable insights, techniques developed in the explainable artificial intelligence (XAI) research community are gaining increasing attention in semiconductor data analytics. One key approach is \textit{Shapley values} (SV)~\cite{roth1988shapley,strumbelj2010efficient,lundberg2017unified}, which compute attribution scores for each input variable in a prediction model, given an observed input-output pair. When applied to a VM model, SV can provide valuable insights into which input variables are most responsible for an observed process quality metric. However, for cross-process defect attribution, existing approaches (e.g.,~\cite{torres2020machine,senoner2022using,lee2023expandable,guo2024enhanced}) have two main limitations: they fail to respect the sequential nature of semiconductor processes and rely on an arbitrarily chosen baseline (or reference) point. These issues stem from the original modeling assumptions of SV.

In this paper, we propose a model-agnostic input attribution algorithm called \textit{Trajectory Shapley Attribution (TSA)}, a novel extension of SV designed for scenarios where input variables form a trajectory. As discussed later, TSA eliminates dependency on an arbitrary baseline point by incorporating a data imputation algorithm that leverages the lot-wafer hierarchy in semiconductor manufacturing. We apply our methodology to in-line measurements from a state-of-the-art front-end-of-line (FEOL) process at the Albany NanoTech fab to demonstrate the utility of our approach.

\section{Related Work}

The overall framework of in-line measurement attribution aligns with advanced process control (APC), which encompasses virtual metrology (VM), variable attribution, and recipe tuning as key sub-tasks. For a recent and comprehensive review of APC, VM, and related tasks, see~\cite{maitra2024virtual}. However, most existing studies focus on specific processes. While cross-process dependency analysis and attribution of wafer quality issues are crucial for accelerating yield improvement~\cite{xu2024fast}, the vision of digital twins---namely, data-driven, semi-automated yield control---remains far from reality. In many mass production fabs, yield ramp-up still requires significant manual effort. 

The integration of model-agnostic XAI techniques with advanced quality prediction models is a major research trend in semiconductor analytics. This approach aims to strike an optimal balance between predictive power and interpretability. Among various XAI methods (see, e.g.,~\cite{molnar2020interpretable} for a general overview), the Shapley value (SV) is the most widely used in semiconductor analytics. Torres et al.~\cite{torres2020machine} and Stoner et al.~\cite{senoner2022using} were among the first to leverage SV for fab-wide yield control. Lee et al.~\cite{lee2023expandable} applied SV to ten different regression models for yield prediction and attribution, demonstrating its model-agnostic nature. Guo et al.~\cite{guo2024enhanced} further utilized SV for chip probing yield in a good-bad classification setting.

Ironically, despite its widespread adoption, most studies treat SV as a \textit{black-box} software tool, paying little attention to the validity of its modeling assumptions. We argue that the original definition of SV may not be fully appropriate for semiconductor manufacturing due to two key issues: 1) variable coalitions include physically inadmissible process routes, and 2) the default baseline value in SHAP, a popular Python implementation~\cite{lundberg2017unified}, disregards the wafer-lot hierarchy. The proposed Trajectory Shapley Attribution (TSA) addresses these limitations.

Finally, the Transformer architecture has recently gained attention in semiconductor analytics~\cite{yella2021soft,han2023deep} due to its ability to directly model dependencies between sequential input variables via the self-attention mechanism~\cite{labaien2023diagnostic}. However, Transformers are known to be even more data-hungry than RNNs. It remains an open question whether Transformers provide reliable predictive performance in semiconductor applications, given the challenge of extremely low effective sample sizes, as discussed in Sec.~\ref{subsec:effective_samples_size}. For instance, Han et al.~\cite{han2023deep} evaluated their proposed Transformer-based VM model using only 344 training samples and 36 validation samples---orders of magnitude smaller than the ``internet-scale'' datasets used in text, speech, and vision domains, where RNN and Transformer architectures have demonstrated remarkable success~\cite{bommasani2021opportunities,zheng2025learning}. Moreover, in the VM setting, Transformer input `tokens' are aggregated in some manner to predict the target variable, introducing an additional explainability challenge.

\section{Problem Setting}

This section outlines the machine learning tasks we address and the key modeling assumptions.

\subsection{Attribution vs. Prediction}

The attribution task considered in this paper is defined as follows:
\begin{definition}[In-line measurement attribution] 
\label{def:attribution}
Given a binary wafer quality label $y \in \{0,1\}$ and a set of in-line measurements $\bmx \in \mathbb{R}^D$ (i.e., a $D$-dimensional real-valued vector), compute the attribution score $s_i(\bmx)$ that quantifies the contribution of each measurement item $i=1,\ldots,D$ to the outcome $y$, serving as a proxy for the underlying process.
\end{definition}
Here, the binary label is defined as 1 for bad wafers and 0 for good wafers. As illustrated in Fig.~\ref{fig:Trajectory_shapley}, each measurement tool is associated with a specific processing tool and acts as its proxy. Some measurements may be performed multiple times on the same wafer due to reentrant processes. In such cases, only the latest observations are retained, allowing the in-line measurement history to be represented as a fixed-dimensional vector $\bmx \in \mathbb{R}^D$ per wafer, although some measurement items may be missing. A single measurement tool can produce multiple measurement items. For example, a set of critical dimensions may be measured at multiple locations on a wafer, with their mean and standard deviation recorded. Consequently, $D$ can be on the order of thousands.

Typically, solving the attribution task requires a predictive model for $y$ as a function of $\bmx$, which we assume unknown in our problem setting. Since $y$ is binary, we consider the following classification problem:
\begin{definition}[Wafer classifier training] 
\label{def:classification}
Given a training dataset $\calD \triangleq \{(\bmx^{(n)}, y^{(n)})\}_{n=1}^N$, train a classifier $p(\bmx)$ that predicts the probability of $y=1$ (bad) given an input $\bmx$.
\end{definition}
Here, $\bmx^{(n)} \in \mathbb{R}^D$ denotes the set of observed in-line measurement items for the $n$-th wafer, and $y^{(n)}$ is its corresponding wafer quality label.

\subsection{Key Data Characteristics} \label{subsec:effective_samples_size}

Although the tasks defined above resemble standard attribution and classification problems, semiconductor manufacturing introduces specific challenges that must be considered:
\begin{enumerate}
    \item Wafers are not necessarily independent due to lot- and batch-based processing.
    \item Wafers are processed sequentially, implying an inherent ordering among measurement tools.
    \item Most in-line measurement values are unavailable due to wafer-, lot-, and batch-wise sampling policies.
\end{enumerate}
The first and second points render many off-the-shelf machine learning tools ineffective, as most assume sample independence and permutation invariance among variables. Regarding the third point, since wafers are randomly selected for measurement at each tool, the resulting measurement trajectories are nearly unique for each wafer. Furthermore, in the yield ramp-up phase, where various processing recipes are tested simultaneously, the diversity of processing routes is potentially large. This further complicates model training, as the \textit{effective sample size}, defined as the number of wafers per unique processing or measurement condition, may be extremely small---possibly as low as one.

To illustrate the characteristics of a typical wafer dataset, we summarize key statistics from the dataset used in our empirical evaluation. As shown in Table~\ref{tab:mea_summary_mea}, the dimensionality $D$ far exceeds the number of wafers, making the prediction task ill-posed. The extremely high missing rate exacerbates this challenge.

\begin{table}[tbh]
    \centering
    \caption{Summary statistics of our dataset collected from the Albany NY CREATES fab.}
    \label{tab:mea_summary_mea}
    \begin{tabular}{cccc}
        \toprule
        $N$ & $D$ & Average missing rate & $\mathrm{Prob}(y=1)$ \\
        \midrule
        572 & $23\,498$ & 94\% & 51\% \\
        \bottomrule
    \end{tabular}
\end{table}

\section{Preliminary: Shapley Values}

Before getting into the details of the proposed approach, this section reviews existing Shapley value (SV)-based attribution methods.

\subsection{Definition of SV}

The goal of Shapley Value (SV) is to evaluate the contribution of each input variable to a model’s prediction. Suppose there are $D$ input variables, denoted as $\bmx \triangleq (x_1,\ldots, x_D)^\top$, and a prediction model $y=f(\bmx)$, which can produce the probability $p(\bmx)$ in the classification task defined in Definition~\ref{def:classification}. Formally, the SV for the $i$-th variable is defined as~\cite{sundararajan2020many,borgonovo2024many}
\begin{align}\label{eq:SV_def}
    \text{SV}_i(\bmx^t) = \frac{1}{D}\sum_{k=1}^D\sum_{\calS \in \calS^{(i)}_k}\binom{D-1}{k-1}^{-1}
    \left[
     v(\calS) - v(\calS-\{i\})
    \right],
\end{align}
where $\bmx^t$ is a test sample at which SV is calculated, and $\calS^{(i)}_k$ denotes the collection of \textit{all} index sets of size $k$ (out of the $D$ members) that include $i$ (Examples are given below). The notation $\calS - \{i\}$ represents the index set obtained by removing $i$ from $\calS$. The dependency on $\bmx^t$ indicates that SV offers a \textit{local} explanation, meaning that the explanation is about the specific test sample $\bmx^t$. This contrasts with traditional correlation analysis, where the relevance to a target variable $y$ is quantified through a statistic defined on a population of samples.

In the original game-theoretic definition~\cite{roth1988shapley}, the term $v(\calS) - v(\calS-\{i\})$ quantifies the influence of the $i$-th player within the coalition $\calS$, assuming that $v(\calS)$, known as the \textit{value function}, represents the outcome achieved by the coalition. Since the number of such possible sets is $\binom{D-1}{k-1}$, SV$_i$ computes the average contribution of the $i$-th variable across all possible variable combinations. While multiple definitions of $v(\cdot)$ are possible in XAI applications~\cite{sundararajan2020many,borgonovo2024many}, it is typically chosen as the prediction function itself, as discussed in the next subsection.

\subsection{Handling Non-Participating Variables}

A critical question when applying SV in XAI tasks is how to reconcile $v(\calS)$, which is defined over a subset of variables, with $f(\bmx)$, which requires a full $D$-dimensional input. To be specific, consider the case where $(i,D)=(i,3)$ and $\calS=\{1,2\}$. In SHAP's default setting, the value function $v(\calS)$ evaluated at $\bmx = \bmx^t$ is defined as
\begin{align}
    v(\{1,2\}) = f(x_1^t,x_2^t,\bar{x}_3), \quad \text{(baseline Shapley)} 
\end{align}
where $\bar{x}_3$ is the population mean of $x_3$. In this setting, the selected variables in $\calS$ retain their actual values from $\bmx^t$, while non-selected variables are replaced with default values, which in this case is the population mean. In other application domains, alternative default values are used; for example, in computer vision, pixel values may be set to zero. These variants are collectively referred to as \textit{baseline Shapley values}. 

Another widely used approach marginalizes the non-selected variables using training data:
\begin{align}
    v(\{1,2\}) = \frac{1}{N}\sum_{n=1}^Nf(x_1^t,x_2^t,x_3^{(n)}),
     \quad \text{(CE Shapley)} 
\end{align}
which is known as the \textit{conditional expectation (CE) Shapley value}~\cite{sundararajan2020many}.

\subsection{Limitations of SV in Semiconductor Applications}

There are two major limitations when applying SV-based attribution using Eq.~\eqref{eq:SV_def} to semiconductor manufacturing data. 

\textit{First}, while SV provides a theoretically fair method for evaluating variable dependencies, it does not account for the sequential nature of semiconductor processes. In the present context, a set $\calS$ can be viewed as a hypothetical processing route, represented by in-line measurements as proxies. However, many variable combinations are physically inadmissible---for instance, pattern formation without etching. It is unclear whether attribution scores derived from such inadmissible combinations carry meaningful information or should be regarded as artifacts.

\textit{Second}, SV relies on an arbitrarily chosen baseline point. The use of the population mean as a reference can be problematic, as it disregards the wafer-lot hierarchy. A similar issue arises with CE Shapley, where averaging over training samples does not necessarily preserve the structure of semiconductor processing dependencies.

Beyond these issues, SV is also known to exhibit \textit{deviation insensitivity}~\cite{ide2023generative} when the prediction function itself is used as the value function. This means that while SV is effective for characterizing the general behavior of $f(\cdot)$, it is not well-suited for explaining deviations of the form $f(\bmx^t) - y^t$, which is often of primary interest in defect diagnosis tasks.

\section{Proposed Approach}

Given the limitations of standard SVs, we propose a novel SV-based framework for measurement attribution. This section introduces two key components of the trajectory Shapley attribution (TSA) framework: lot-aware kernel imputation and trajectory Shapley values.

\subsection{Lot-Aware Kernel Imputation}

As discussed earlier, in-line measurement data contain many missing entries due to lot-, wafer-, and item-specific sampling policies. A key challenge is that missing patterns in in-line measurements are neither purely random nor fully systematic. In the TSA framework, we adopt a relatively simple yet effective data imputation approach based on the lot-wafer hierarchy. This imputation method is crucial, as it defines the concept of `non-participation' for variables.

Let $W_{n,m}$ denote the kernel function between wafers $n$ and $m$. To compute $W$, we first define the lot-sharing count matrix $C$, where each element $C_{n,m}$ represents the number of operations shared between wafers $n$ and $m$ within the same lot, up to the point where either $y^{(n)}$ or $y^{(m)}$ is measured. $C_{n,n}$ corresponds to the total number of operations that wafer $n$ underwent before the measurement of $y^{(n)}$. Using $C$, we define the similarity matrix $W$ based on the Jaccard index:
\begin{align}
W_{n,m} \triangleq 
        \frac{C_{n,m}}{C_{m,m}+C_{n,n}-C_{n,m}}, \quad  n\neq m.
\end{align}
Each element of $W$ lies within the unit interval $[0, 1]$, with diagonal elements $W_{n,n}$ set to zero.

Finally, LAKI imputes a missing element, denoted as $x_k^{(n)}=\bot$, using a weighted average based on $W$:
\begin{align}\label{eq:LAKI_imputation}
    x_{k}^{(n)} \leftarrow 
         \frac{1}{\sum_{j } W_{n,j}} \sum_{m} W_{n,m} x_{k}^{(m)},
\end{align}
where terms with $x_{k}^{(m)} = \bot$ are excluded from the summation. If no peers with observed values for item $k$ exist, we set $x_k^{(n)}$ to $\bar{x}_k$, the mean of item $k$.

\subsection{Trajectory Shapley Values}

Suppose we have a VM model $f(\bmx)$ that predicts an outcome variable $y$ given a measurement trajectory $\bmx$. We generalize the standard SV definition~\eqref{eq:SV_def} as follows:
\begin{align}\label{eq:traj_SV}
    s_i(\bmx^t) &= \sum_{\calS \in \mathcal{T}^{(i)} } \mu(\calS) 
    \left[v(\calS) - v(\calS- \{i\}) \right],
\end{align}
where $\bmx^t$ denotes an input trajectory of interest, and $\calT^{(i)}$ is the set of all \textit{physically admissible} process trajectories that include the $i$-th measurement item. In semiconductor manufacturing, physically admissible trajectories are those that progress sequentially from the first process without skipping intermediate steps. For example, for $(i, D)=(2,3)$, the admissible trajectories are $(x_1,x_2)$ and $(x_1, x_2, x_3)$. The prefactor $\mu(\calS)$ is a weighting function that averages over all admissible configurations. To properly determine $\mu(\calS)$, we impose the sum rule on the trajectory Shapley values:
\begin{align}\label{eq:traj_SV_sumrule}
    \sum_{i=1}^D s_i(\bmx^t) &= v(\bmone) - v(\emptyset),
\end{align}
where $v(\bmone)$ and $v(\emptyset)$ denote the value function under full variable participation and no participation, respectively. Equations~\eqref{eq:traj_SV} and~\eqref{eq:traj_SV_sumrule} together define the trajectory Shapley value within our framework.

To define the value function, we relabel the index set $\calS$ and its complement $-\calS$ so that $\bmx = (\bmx_{\calS}, \bmx_{-\calS})$, where the subvector $\bmx_{\calS}$ collects all $x_i$'s for $i \in \calS$. The value function in TSA, evaluated at $\bmx=\bmx^t$, is then given by:
\begin{align}\label{eq:TSA_value_function_def}
    v(\calS) = f(\bmx_{\calS}=\bmx^t, \bmx_{-\calS}= \perp), \quad \text{(TSA)}
\end{align}
where $\bmx_{-\calS} = \perp$ symbolically represents that the non-participating variables are treated as missing values. Unlike standard SV, where missing variables are replaced with a fixed baseline value, TSA employs LAKI to adaptively impute $\perp$ based on the wafer and lot IDs.

Our theoretical analysis has established the following result:
\begin{theorem}[Trajectory Shapley Value]
    The weighting function $\mu(\calS)$ is uniquely determined by the condition~\eqref{eq:traj_SV_sumrule}. With this $\mu(\calS)$, the trajectory Shapley value is given by:
    \begin{align}\label{eq:Traj_SV_intervention}
     s_i(\bmx^t) = f(\bmx_{1:i}^t,\bmx_{(i+1):D}^0) - f(\bmx_{1:(i-1)}^t,\bmx_{i:D}^0),
\end{align}
where $\bmx_{s:e} \triangleq (x_s, x_{s+1}, \dots, x_e)$ for $s\leq e$, and $\emptyset$ (the empty set) otherwise. $\bmx_{(i+1):D}^0$ and $\bmx_{i:D}^0$ are baseline values computed using the LAKI algorithm~\eqref{eq:LAKI_imputation}. 
\end{theorem}
The proof is omitted due to space constraints. In the next section, we apply this attribution score to a good-bad wafer characterization scenario.

\section{Empirical Evaluation}

\begin{figure}[tb]
    \centering
    \includegraphics[trim = 60mm 25mm 60mm 35mm, clip,width=1\linewidth]{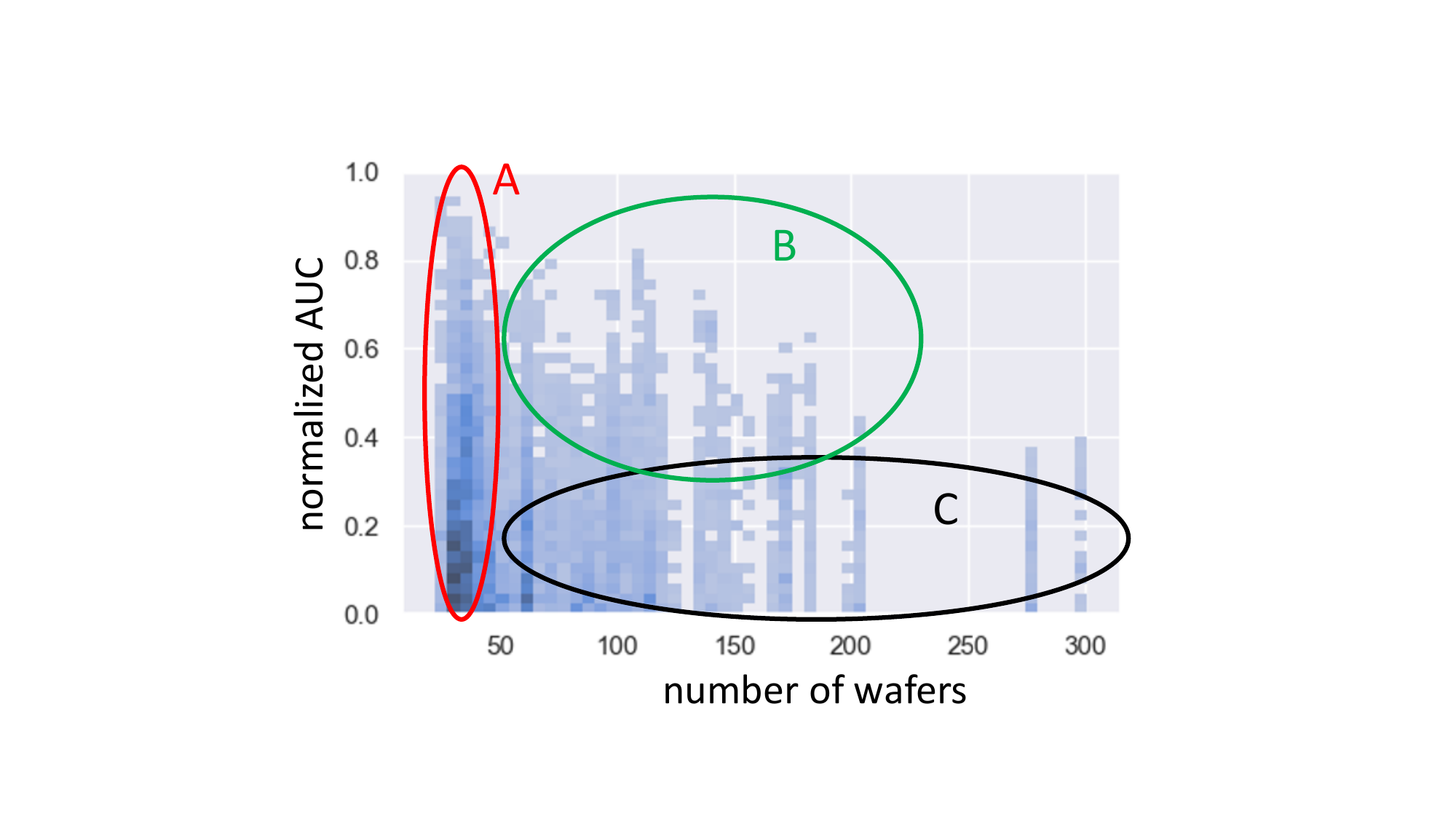}
    \caption{Results of univariate correlation analysis. Most measurement items either provide weak signals or have too few samples, highlighting the need for trajectory-based analysis.}
    \label{fig:Univariate}
\end{figure}


We applied the Trajectory Shapley Attribution (TSA) approach to a process-limited yield (PLY) root-cause analysis task for state-of-the-art front-end-of-line (FEOL) processing routes in the Albany NanoTech fab. In this empirical evaluation, we focused on a specific scenario involving a burst of a defect type over a certain period. Specifically, we collected data from $N=572$ wafers, approximately half of which exhibited the defect. A binary target variable $y \in \{0,1\}$ was obtained at a PLY measurement point, where $y=1$ indicates the presence of a specific defect type, and $y=0$ otherwise. The input variables $\bmx$ consisted of upstream in-line measurements from hundreds of measurement processes. 

As shown in Table~\ref{tab:mea_summary_mea}, the missing value rate is extremely high---only about 6\% of the entries contain actual measurements. Additionally, the dimensionality of the input space ($D$) is much larger than the number of wafers ($N$), making the prediction task highly ill-posed. In such settings, data-hungry deep learning models are prone to overfitting. Our goal is to identify which measurement items are most relevant to the defect burst by computing attribution scores.

\subsection{Baseline: Conventional Univariate Analysis}

As a baseline approach, we conducted a conventional univariate correlation analysis. Using the training dataset $\calD$, we trained a separate univariate logistic regression classifier for each of the $D= 23\,498$ measurement items to predict the good-bad wafer flag $y$. No LAKI imputation was used in this step. Due to missing values, different measurement items had varying sample sizes, which impacted prediction accuracy. For each measurement item, we computed the area under the curve (AUC) score of the receiver operating characteristic (ROC) curve.

Figure~\ref{fig:Univariate} presents the results, where the vertical axis represents the normalized AUC, defined as $|2\times \text{AUC} -1|$, which takes a value of 0 for the random classifier and 1 for the perfect classifier. As shown, most measurement items produced unreliable results due to insufficient sample sizes (Region A). Generally, measurement items with larger sample sizes exhibited weaker predictive performance (Region C). Although a few items displayed moderate predictive power, no clear dominant predictor emerged (Region B), providing limited actionable insights.

\subsection{Training a Good-Bad Classifier}

As the first step in the TSA analysis, we trained a binary classifier in the multivariate setting. To mitigate potential overfitting issues (see Sec.~\ref{subsec:effective_samples_size}), we employed $\ell_2$-regularized logistic regression to model $p(\bmx)$, the probability of observing $y=1$ (bad wafer). The regularization strength was selected via cross-validation. With LAKI imputation, the true positive rate was 0.80, compared to 0.75 without LAKI, demonstrating the effectiveness of the LAKI strategy.

\subsection{Results of Trajectory Shapley Attribution}

Using the trained classifier $p(\bmx)$ and setting $f(\bmx)\triangleq p(\bmx)$, we computed TSA attribution scores for each wafer outside the training dataset with Eq.~\eqref{eq:Traj_SV_intervention}:
\begin{align}\label{eq:Traj_SV_intervention2}
     s_i(\bmx^t) = p(\bmx_{1:i}^t,\bmx_{(i+1):D}^0) - p(\bmx_{1:(i-1)}^t,\bmx_{i:D}^0).
\end{align}

\begin{figure}[tbp]
    \centering
    \includegraphics[clip, trim=0.5cm 1.6cm 0.2cm 5cm, width=0.49\textwidth]{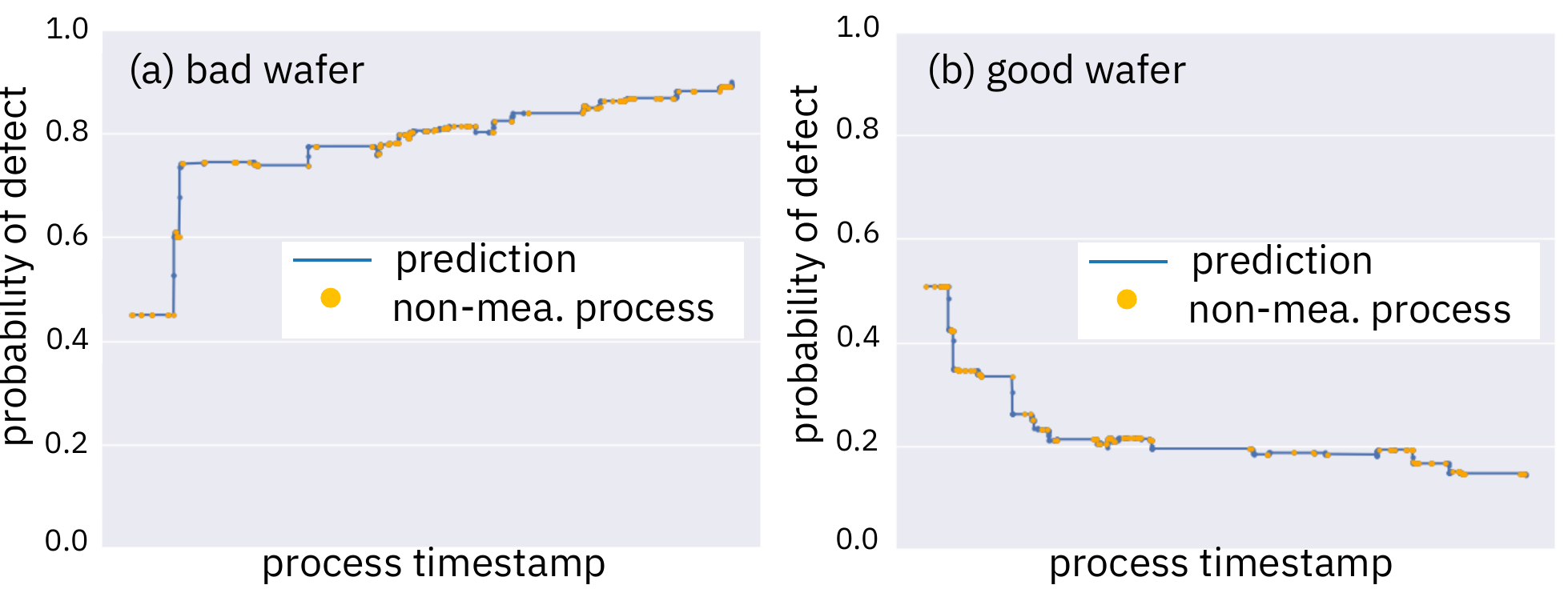}
    \vspace{-5mm}
    \caption{
        Cumulative attribution score, $\beta(\tau,\bmx^t)$, defined in Eq.~\eqref{eq:cummulative_TSA}, plotted against process timestamp $\tau$. $\bmx^t$ is the vector of inline measurements of the specified wafer. The plot shows how the baseline prediction $p(\bmx^0)$ evolves into the final prediction $p(\bmx^t)$ as time progress.
    }
    \label{fig:scoring}
\end{figure}

An effective way to assess the \textit{relative} importance of each measurement item is to visualize the scores using a cumulative attribution plot.  Figure~\ref{fig:scoring} shows two examples of such cross-process TSA visualizations. By leveraging the sum rule~\eqref{eq:traj_SV_sumrule}, we can visualize how `badness' accumulates as a wafer progresses through the processing route. 

Specifically, we plot the \textit{cumulative attribution score} up to time $\tau$, defined as:
\begin{align}\label{eq:cummulative_TSA}
    \beta(\tau,\bmx^t) \triangleq v(\emptyset) + \sum_{k=1}^D s_k(\bmx^t) \mathbb{I}(t_k \leq \tau),
\end{align}
where $t_k$ is the timestamp of the $k$-th measurement item and $\mathbb{I}(\cdot)$ is the indicator function that equals 1 if the condition in the parenthesis is satisfied, and 0 otherwise. For the specific expression in Eq.~\eqref{eq:Traj_SV_intervention2}, the sum rule leads to
\begin{align}
    \beta(0,\bmx^t) = p(\bmx^0), \quad
    \beta(\infty,\bmx^t) = p(\bmx^t),
\end{align} 
assuming that the process starts after $\tau=0$. These equations determine the start and end points of the cumulative attribution plot in Fig.~\ref{fig:scoring}.

Notably, certain time points exhibit significant jumps, suggesting potential root causes for favorable or unfavorable outcomes. In particular, Figure~\ref{fig:scoring}(a) indicates that our TSA approach could have facilitated earlier process termination for defective wafers, potentially months before reaching the PLY measurement point.

\section{Conclusion}

We have proposed the Trajectory Shapley Attribution (TSA) method as a model-agnostic and interpretable framework for root-cause analysis in sequential manufacturing processes.

The proposed method addresses two major limitations of standard Shapley value (SV)-based attribution: reliance on unrealistic process routes and sensitivity to an arbitrary baseline point. TSA resolves these issues by introducing two novel components---lot-aware kernel imputation and trajectory Shapley values. We mathematically demonstrated that the trajectory Shapley value reduces to a closed-form expression without combinatorial complexity when the sum rule is imposed.

We demonstrated the practical utility of TSA through a case study in a good-bad wafer classification task, using real-world data from the Albany NanoTech fab.

While TSA was developed as a variant of the Shapley value, other algorithmic approaches may also be effective for attribution in semiconductor manufacturing. Exploring alternative methods, such as partial trajectory regression~\cite{Miyaguchi25ASMC2}, and establishing a unified framework for defect root-cause analysis represent promising directions for future research.


\section*{Acknowledgement}

The authors gratefully acknowledge the support of NY CREATES and the Albany NanoTech Complex for providing access to state-of-the-art fabrication and characterization resources. They also extend their gratitude to Dr.~Monirul Islam and Dr.~Ishtiaq Ahsan for providing the PLY data and their valuable support throughout the project.

\bibliographystyle{ieeetr}
\bibliography{references}

\begin{thebibliography}{10}

\bibitem{susto2015multi}
G.~A. Susto, S.~Pampuri, A.~Schirru, A.~Beghi, and G.~De~Nicolao, ``Multi-step virtual metrology for semiconductor manufacturing: A multilevel and regularization methods-based approach,'' {\em Computers \& Operations Research}, vol.~53, pp.~328--337, 2015.

\bibitem{jebri2016virtual}
M.~A. Jebri, E.~El~Adel, G.~Graton, M.~Ouladsine, and J.~Pinaton, ``Virtual metrology on semiconductor manufacturing based on just-in-time learning,'' {\em IFAC-PapersOnLine}, vol.~49, no.~12, pp.~89--94, 2016.

\bibitem{kim2018variable}
K.-J. Kim, K.-J. Kim, C.-H. Jun, I.-G. Chong, and G.-Y. Song, ``Variable selection under missing values and unlabeled data in semiconductor processes,'' {\em IEEE Transactions on Semiconductor Manufacturing}, vol.~32, no.~1, pp.~121--128, 2018.

\bibitem{yella2021soft}
J.~Yella, C.~Zhang, S.~Petrov, Y.~Huang, X.~Qian, A.~A. Minai, and S.~Bom, ``Soft-sensing conformer: A curriculum learning-based convolutional transformer,'' in {\em 2021 IEEE International Conference on Big Data (Big Data)}, pp.~1990--1998, IEEE, 2021.

\bibitem{han2023deep}
S.~{Han, et al.}, ``Deep learning-based virtual metrology in multivariate time series,'' in {\em 2023 IEEE International Conference on Prognostics and Health Management (ICPHM)}, pp.~30--37, IEEE, 2023.

\bibitem{dalla2023deep}
F.~Dalla~Zuanna, N.~Gentner, and G.~A. Susto, ``Deep learning-based sequence modeling for advanced process control in semiconductor manufacturing,'' {\em IFAC-PapersOnLine}, vol.~56, no.~2, pp.~8744--8751, 2023.

\bibitem{lee2020recurrent}
K.~B. Lee and C.~O. Kim, ``Recurrent feature-incorporated convolutional neural network for virtual metrology of the chemical mechanical planarization process,'' {\em Journal of Intelligent Manufacturing}, vol.~31, no.~1, pp.~73--86, 2020.

\bibitem{hsu2023virtual}
C.-Y. Hsu and Y.-W. Lu, ``Virtual metrology of material removal rate using a one-dimensional convolutional neural network-based bidirectional long short-term memory network with attention,'' {\em Computers \& Industrial Engineering}, vol.~186, p.~109701, 2023.

\bibitem{roth1988shapley}
A.~E. Roth, {\em The Shapley value: essays in honor of Lloyd S. Shapley}.
\newblock Cambridge University Press, 1988.

\bibitem{strumbelj2010efficient}
E.~Strumbelj and I.~Kononenko, ``An efficient explanation of individual classifications using game theory,'' {\em The Journal of Machine Learning Research}, vol.~11, pp.~1--18, 2010.

\bibitem{lundberg2017unified}
S.~M. Lundberg and S.-I. Lee, ``A unified approach to interpreting model predictions,'' in {\em Proceedings of the 31st International Conference on Neural Information Processing Systems}, NIPS'17, p.~4768–4777, 2017.

\bibitem{torres2020machine}
J.~A. Torres, I.~Kissiov, M.~Essam, C.~Hartig, R.~Gardner, K.~Jantzen, S.~Schueler, and M.~Niehoff, ``Machine learning assisted new product setup,'' in {\em 2020 31st Annual SEMI Advanced Semiconductor Manufacturing Conference (ASMC)}, pp.~1--5, IEEE, 2020.

\bibitem{senoner2022using}
J.~Senoner, T.~Netland, and S.~Feuerriegel, ``Using explainable artificial intelligence to improve process quality: evidence from semiconductor manufacturing,'' {\em Management Science}, vol.~68, no.~8, pp.~5704--5723, 2022.

\bibitem{lee2023expandable}
Y.~Lee and Y.~Roh, ``An expandable yield prediction framework using explainable artificial intelligence for semiconductor manufacturing,'' {\em Applied Sciences}, vol.~13, no.~4, p.~2660, 2023.

\bibitem{guo2024enhanced}
P.~Guo and Y.~Chen, ``Enhanced yield prediction in semiconductor manufacturing: Innovative strategies for imbalanced sample management and root cause analysis,'' in {\em 2024 IEEE International Symposium on the Physical and Failure Analysis of Integrated Circuits (IPFA)}, pp.~1--6, IEEE, 2024.

\bibitem{maitra2024virtual}
V.~Maitra, Y.~Su, and J.~Shi, ``Virtual metrology in semiconductor manufacturing: Current status and future prospects,'' {\em Expert Systems with Applications}, p.~123559, 2024.

\bibitem{xu2024fast}
H.-W. Xu, Q.-H. Zhang, Y.-N. Sun, Q.-L. Chen, W.~Qin, Y.-L. Lv, and J.~Zhang, ``A fast ramp-up framework for wafer yield improvement in semiconductor manufacturing systems,'' {\em Journal of Manufacturing Systems}, vol.~76, pp.~222--233, 2024.

\bibitem{molnar2020interpretable}
C.~Molnar, {\em Interpretable machine learning}.
\newblock Lulu.com, 2020.

\bibitem{labaien2023diagnostic}
J.~Labaien, T.~Id{\'e}, P.-Y. Chen, E.~Zugasti, and X.~De~Carlos, ``Diagnostic spatio-temporal transformer with faithful encoding,'' {\em Knowledge-Based Systems}, vol.~274, p.~110639, 2023.

\bibitem{bommasani2021opportunities}
R.~Bommasani, D.~A. Hudson, E.~Adeli, R.~Altman, S.~Arora, S.~von Arx, M.~S. Bernstein, J.~Bohg, A.~Bosselut, E.~Brunskill, {\em et~al.}, ``On the opportunities and risks of foundation models,'' {\em arXiv preprint arXiv:2108.07258}, 2021.

\bibitem{zheng2025learning}
H.~Zheng, L.~Shen, A.~Tang, Y.~Luo, H.~Hu, B.~Du, Y.~Wen, and D.~Tao, ``Learning from models beyond fine-tuning,'' {\em Nature Machine Intelligence}, pp.~1--12, 2025.

\bibitem{sundararajan2020many}
M.~Sundararajan and A.~Najmi, ``The many shapley values for model explanation,'' in {\em International conference on machine learning}, pp.~9269--9278, PMLR, 2020.

\bibitem{borgonovo2024many}
E.~Borgonovo, E.~Plischke, and G.~Rabitti, ``The many shapley values for explainable artificial intelligence: A sensitivity analysis perspective,'' {\em European Journal of Operational Research}, 2024.

\bibitem{ide2023generative}
T.~Id{\'e} and N.~Abe, ``Generative perturbation analysis for probabilistic black-box anomaly attribution,'' in {\em Proceedings of the 29th ACM SIGKDD Conference on Knowledge Discovery and Data Mining}, pp.~845--856, 2023.

\bibitem{Miyaguchi25ASMC2}
K.~Miyaguchi, M.~Joko, R.~Sheraw, and T.~Id\'{e}, ``Wafer defect root cause analysis with partial trajectory regression,'' in {\em 2025 SEMI Advanced Semiconductor Manufacturing Conference (ASMC)}, p.~TBD, IEEE, 2025.

\end{thebibliography}

\end{document}